\documentclass{article} % For LaTeX2e
\usepackage{iclr2025_conference,times}

\usepackage{amsmath,amsfonts,bm}

\def\eqref#1{equation~\ref{#1}}
\def\1{\bm{1}}

\DeclareMathAlphabet{\mathsfit}{\encodingdefault}{\sfdefault}{m}{sl}
\SetMathAlphabet{\mathsfit}{bold}{\encodingdefault}{\sfdefault}{bx}{n}

\usepackage{tabularx}
\usepackage{graphicx}
\usepackage{svg}
\usepackage{booktabs}
\usepackage{pdfpages}
\usepackage{pax}
\usepackage{pgfplots}
\pgfplotsset{compat=1.18}
\pgfplotsset{every tick label/.append style={font=\scriptsize}}
\usepackage{subcaption}
\usepackage{hyperref}
\usepackage{cleveref}
\usepackage{url}

\title{Representing Signs as Signs: One-Shot ISLR to Facilitate Functional Sign Language Technologies}

\author{Toon Vandendriessche, Mathieu De Coster, Annelies Lejon \& Joni Dambre \\
IDLab-AIRO -- Ghent University -- imec\\
Ghent, Belgium \\
\texttt{\{firstname.lastname\}@ugent.be} \\
}

\newcommand{\PFASL}{PF-ASL}%{PoseFormer\textsubscript{ASL}}
\newcommand{\PFVGT}{PF-VGT}%{PoseFormer\textsubscript{VGT}}

\iclrfinalcopy % Uncomment for camera-ready version, but NOT for submission.
\begin{document}

\maketitle

\begin{abstract}
% \mathieu{mooi abstract!}
    Isolated Sign Language Recognition (ISLR) is crucial for scalable sign language technology, yet language-specific approaches limit current models. To address this, we propose a one-shot learning approach that generalises across languages and evolving vocabularies. Our method involves pretraining a model to embed signs based on essential features and using a dense vector search for rapid, accurate recognition of unseen signs. We achieve state-of-the-art results, including 50.8\% one-shot MRR on a large dictionary containing 10,235 unique signs from a different language than the training set. Our approach is robust across languages and support sets, offering a scalable, adaptable solution for ISLR. Co-created with the Deaf and Hard of Hearing (DHH) community, this method aligns with real-world needs, and advances scalable sign language recognition.
  % This paper presents a new baseline for a simple, yet highly effective, method for one-shot isolated sign language recognition. The simplicity of the method stems from the fact that the one-shot classifier is a frozen neural network and only requires a trivial initialisation but no weight updates through training. Yet, at the same time, the method is highly effective, as we achieve high recall despite having only one training example per sign. We show that, with a sufficiently large and varied pretraining set, it is possible to classify unseen signs for which only one training example is available. This is possible even if the sign originates from a different sign language than the language used for pretraining. This method enables search functionality for online sign language dictionaries, as well as large-vocabulary sign language recognition for sign languages for which no extensive datasets are available.
  % \keywords{Sign language recognition \and One-shot classification \and Human pose estimation}
\end{abstract}

\section{Introduction}
\label{sec:intro}

Isolated sign language recognition (ISLR) is a crucial first step toward achieving full sign language translation (SLT). Despite notable progress, most advances in ISLR have been confined to specific languages and respective datasets, limiting the range of vocabularies that can be recognised in real-world applications. In addition, with sign languages constantly evolving, this limitation becomes increasingly problematic. More flexible approaches are required -- ones that allow for the expansion of vocabularies and are robust enough to handle large-scale dictionaries for SLT.

To overcome these challenges, it is essential to shift the focus away from predefined vocabularies and instead emphasise the intrinsic features of signs. Once an \textit{effective} representation of a specific sign is established, the surrounding context can be integrated afterwards. An \textit{effective} sign representation requires two key elements: (1) each sign is uniquely identifiable, and (2) only the essential features, i.e. sign phonemes, are captured to ensure reliable discrimination and adaptability across varied signing conditions.
% \mathieu{to ensure clarity and precision: dit is nietszeggend vind ik, zou ik weglaten of vervangen door iets concreters}

Following this idea, we propose leveraging one-shot learning for sign language recognition. One-shot learning is particularly well-suited for sign language recognition since it allows the model to generalise from limited examples, making it possible to recognise new signs without extensive retraining or data collection. By focusing on embedding signs efficiently, this method enables accurate, context- and even language-independent isolated sign recognition.

Our approach consists of two steps: (1) pretraining a model that can reliably embed signs, and (2) using this frozen pretrained model for a dense vector search. In the first step, we train the model on a diverse set of signs to capture their essential features. This creates embeddings representing signs as points in a high-dimensional space. In the second step, a dense vector search matches new, unseen variations of signs to the closest embedding in this space, enabling rapid and accurate recognition without requiring further retraining.

We evaluate both the pretraining and one-shot classification, achieving state-of-the-art results. Notably, we attain a 50.8\% one-shot MRR on an extremely large dictionary from a language entirely different from the pretraining data. Our method also demonstrates strong robustness across multiple languages, even when introducing variations in the set of exemplary signs. These findings underscore that ISLR systems can be generalised across languages, independent of the pretraining data. This enables recognition of vast vocabularies while adapting to the evolving nature of sign languages. This adaptability positions our approach as a progressive solution for enhancing the scalability and effectiveness of sign language technology in diverse contexts.
% \mathieu{die laatste zin vind ik er lichtjes over en ruikt een beetje naar generative AI (sorry als dat niet zo is :p) vind ik, persoonlijk zou ik die weglaten}

The development of this paper was guided by a clear message from the DHH community: sign language technology can be initially suboptimal but should be improvable through active feedback, to provide more usable tools for the DHH community \emph{in the short term}. This goal was emphasised at a workshop on sign language research in AI \citep{bragg2019sign}. Bearing this in mind, the method of this paper was developed in a co-creation strategy, where the needs and desires of the DHH community were assessed frequently. Ultimately, this led to an openly available tool\footnote{\url{https://woordenboek.vlaamsegebarentaal.be/signBuddy}}.

% \textbf{[AUTHORS: In the light of this paper, we integrated a publically available application into an online sign language dictionary. This application allows the user to search for a sign in the dictionary by simply performing it in front of a webcam. We can not provide any more details, since this conflicts with the double-blind review policy of ICLR. This paragraph is a placeholder for a more detailed description after publication.]}
% \mathieu{is dit oke volgens de ICLR guidelines om er zo'n paragraaf in te zetten? Plus het wordt vrij obvious dat het over VGT gaat omdat dat de enige dictionary is die besproken wordt in deze paper} \toon{paragraaf wegdoen, zinnetje toevoegen aan bovenstaande en voetnoot met link.}

\textbf{The contributions of this paper are the following.} 
\textbf{(a)} This paper presents the first robust approach to one-shot sign language recognition, demonstrating its language-independent capabilities and its applicability to larger vocabularies.
\textbf{(b)} We achieve state-of-the-art results in sign language recognition.
\textbf{(c)} Our method demonstrates strong generalisation across multiple languages and proves robust even when variations are introduced in the exemplary signs.
\textbf{(d)} We developed an application using a co-creation strategy in close collaboration with the DHH community, ensuring that sign language research output meets their needs and can be continuously improved through active feedback.

\section{Related Work}

\subsection{Isolated Sign Language Recognition}
\label{sec:islr}

ISLR is a classification problem. Traditionally, a system analyses a video depicting an isolated sign and aims to predict the corresponding label or gloss. These input videos can be processed in several ways. Recent advancements \citep{papadimitriou2023sign, chen2022two} have demonstrated a positive impact using pose estimation models, such as MediaPipe Holistic \citep{grishchenko2020mediapipe} (henceforth MediaPipe) and OpenPose \citep{cao2017realtime}. These models transform input videos into sequences of skeletal representations, capturing ``keypoints'' or ``landmarks'' of the human pose in 2D or 3D Cartesian coordinates. By removing all information about a person's appearance, these tools enhance the generalisability to downstream tasks. This transformation allows ISLR models to focus solely on the structural aspect of sign videos: how people move their arms, hands, and face.

% \mathieu{Volgende paragraaf begint alsof het over MediaPipe's voordelen gaat gaan, maar is dan
% terug algemeen. Ik zou de eerste zin ofwel naar boven verplaatsen, ofwel de paragraaf meer over
% deze ene tool laten gaan en niet in het algemeen spreken. Kaggle was ten slotte ook MediaPipe}
MediaPipe has significantly advanced the current state-of-the-art of ISLR, but there is still considerable room for improvement. \citet{moryossef2021evaluating} argued that this tool is not directly applicable to fine-grained tasks like sign language recognition. Although the keypoint estimator is generally accurate, MediaPipe struggles when two body parts interact. Since this interaction is elemental to sign language, crucial information is lost. However, recent Kaggle competitions \citep{asl-signs, chow2023google} based on keypoint estimation using MediaPipe present a different perspective, showing promising results in SLR using keypoint estimation. A key component appears to be the addition of a frame embedding that is not present in the work by Moryossef et al., but present in all top Kaggle competition entries. This frame embedding allows the network to learn the non-linear relationships between keypoints \citep{de2023towards}.

% Overall, MediaPipe provides the optimal balance between accuracy and computational efficiency \cite{de2023towards}, while it also provides the best estimation of the hands. 

% \mathieu{Misschien hier nog vermelden: a key component appears to be the addition frame embedding that is not present in the work by Moryossef et al., but in all top Kaggle competition entries. This frame embedding allows the network to learn
% the non-linear relationships between keypoints [6]. En ook dat MediaPipe beter is dan OpenPose voor handen, bv eerste zin van deze paragraaf naar hier verplaatsen}

% \mathieu{De eerste zin mag concreter. Niet "there are many decisions to be made", maar je mag gewoon zeggen dat de architectuur een grote impact heeft.}
Not only the preprocessing of sign language videos is essential. The architecture of the models also has a great impact. Until 2020, deep learning approaches to ISLR
% \mathieu{Initial vind ik niet ideaal als woord. Misschien "deep learning approaches until 2020"} deep learning approaches 
primarily used variations of Recurrent Neural Networks \citep{koller2016deep, koller2017re, ye2018recognizing}. The common factor of these models is that they are proficient at handling sequential data and dealing with temporal dependencies between different poses. The introduction of transformers \citep{vaswani2017attention} in 2017 initiated a paradigm shift. 
% The first applications of transformers in the domain of ISLR and sign language translation were proposed in 2020: \cite{de2020sign}, respectively \cite{camgoz2020sign}. 
% \mathieu{Misschien hier ook verwijzen naar Sign Language Transformers van Camgoz voor translation, anders zitten we echt wel praktisch enkel mij te citeren en wordt het obvious :p} 
% In the competitions mentioned above, models employing attention mechanisms also performed significantly better.
The combination of keypoints and attention leads to powerful models for ISLR: the top scoring on the Kaggle ASL ISLR competition \citep{asl-signs} method achieved 89.3\% test set accuracy on 250 sign categories using keypoint data.

\subsection{Few-Shot Sign Language Recognition}

% Sign languages are natural languages, and as such, they evolve. This evolution necessitates the development of flexible techniques for ISLR. 
The evolving nature of sign languages motivates the need for flexible ISLR techniques.
The traditional 
% \mathieu{Leuke intro, maar misschien hier kort zeggen wat je bedoelt met traditional (verwijzen naar vorige sectie)} 
sign language classification models described in the previous section lack this flexibility, as they are limited to the glossary provided in the training set and require a large number of examples for every sign. 
% To address this limitation many approaches resort to one- or few-shot learning.\mathieu{De vorige zin "to adress" zou ik weglaten, de volgende is een betere introductie tot dit concept. Nu komt het uit het niets, terwijl de volgende zin de context geeft waarom few of one shot relevant is} 
\citet{fei2006one} argued that ``one can take advantage of knowledge coming from previously learned categories, no matter how different these
categories might be.''  This insight led to the introduction of few-shot learning, where fewer examples per category are required.

There are various gradations to few-shot learning, including zero-shot learning.
%In this approach, the model has never seen an example of the task at hand and completely relies on its prior knowledge.
In zero-shot learning, the model has never seen an example for a given category and relies on its
prior knowledge and some form of description of the category.
Zero-shot learning was first introduced to SLR by \citet{bilge2019zero}. Their method involved matching textual descriptions to video inputs, utilising BERT text embeddings \citep{devlin2018bert} and 3D CNNs alongside bidirectional LSTMs for video processing. More recently, \citet{rastgoo2021zs} approached zero-shot ISLR with a similar technique. This method combined BERT text embeddings with pose estimators and vision transformers, followed by LSTMs for temporal modelling. These zero-shot methods rely on the presence of a textual description of a sign, which is not always available. 

% For languages like VGT a dictionary is often available, a collection of every recognised sign and one corresponding video. 
More often, there are one or multiple example videos available for one sign, which can be used for few-shot learning. If none are available for a given sign, recording one example is easier than accurately describing the sign through text.
% While zero-shot learning offers significant advantages\mathieu{welke advantages? komt niet echt naar boven uit het vorige. Ik zou zeggen dat het uitdagend is, en dat je bv textual descriptions nodig hebt wat je ook niet voor elk gebaar wil doen. Een, of een klein aantal gebaren, verzamelen, kan makkelijker zijn en schaalbaarder. Dus kijken we naar one shot}
 % there are cases where some examples can be beneficial. 
\citet{wang2021cornerstone}, for example, leverage multiple examples (i.e., they perform few-shot learning) of one sign to perform K-means clustering and a custom matching algorithm. For some sign languages, a dictionary is available. In the case of VGT, the dictionary contains exactly one example per unique sign, which is ideal for one-shot learning. \citet{de2023querying}  in essence performed one-shot learning to recognise signs in the VGT dictionary \citep{van2004woordenboek}. In their work, a pretrained model outputs embeddings, which are used in a Euclidean distance-based vector search. However, despite being pretrained on VGT data, the model’s performance on dictionary lookup was suboptimal. This shortfall can be attributed to the Zipfian class distribution in the dataset and the fact that the examples come from continuous signing rather than isolated signs.

\section{Methodology}
\subsection{One-Shot Sign Language Recognition}
\label{sec:oneshot}
% \mathieu{als je de vergelijking met mijn paper nog minder obvious wilt maken, zou je kunnen zeggen dat dictionary search 1 applicatie is van one-shot ISLR (en er hier nog over zwijgen maar dit vermelden op punt waar ik een comment geplaatst heb met note HIERDICTIONARY}

% We adapt the two-stage method proposed by De Coster and Dambre \cite{de2023querying}, to perform one-shot sign language recognition. This approach involves an initial supervised pretraining phase, where the model learns to produce internal sign representations or embeddings. These embeddings are then utilised in the one-shot inference phase. While De Coster and Dambre employed Euclidean distance-based one-nearest-neighbour search for this purpose, we propose instead to leverage matching networks, described by Vinyals et al. \cite{vinyals2016matching}. 
% This method, illustrated in \cref{fig:matching}, offers an alternative approach that provides more interpretable results and a better fit for downstream tasks, all without compromising accuracy.

\begin{figure}[ht!]
    \centering
    \includegraphics[width=0.9\textwidth]{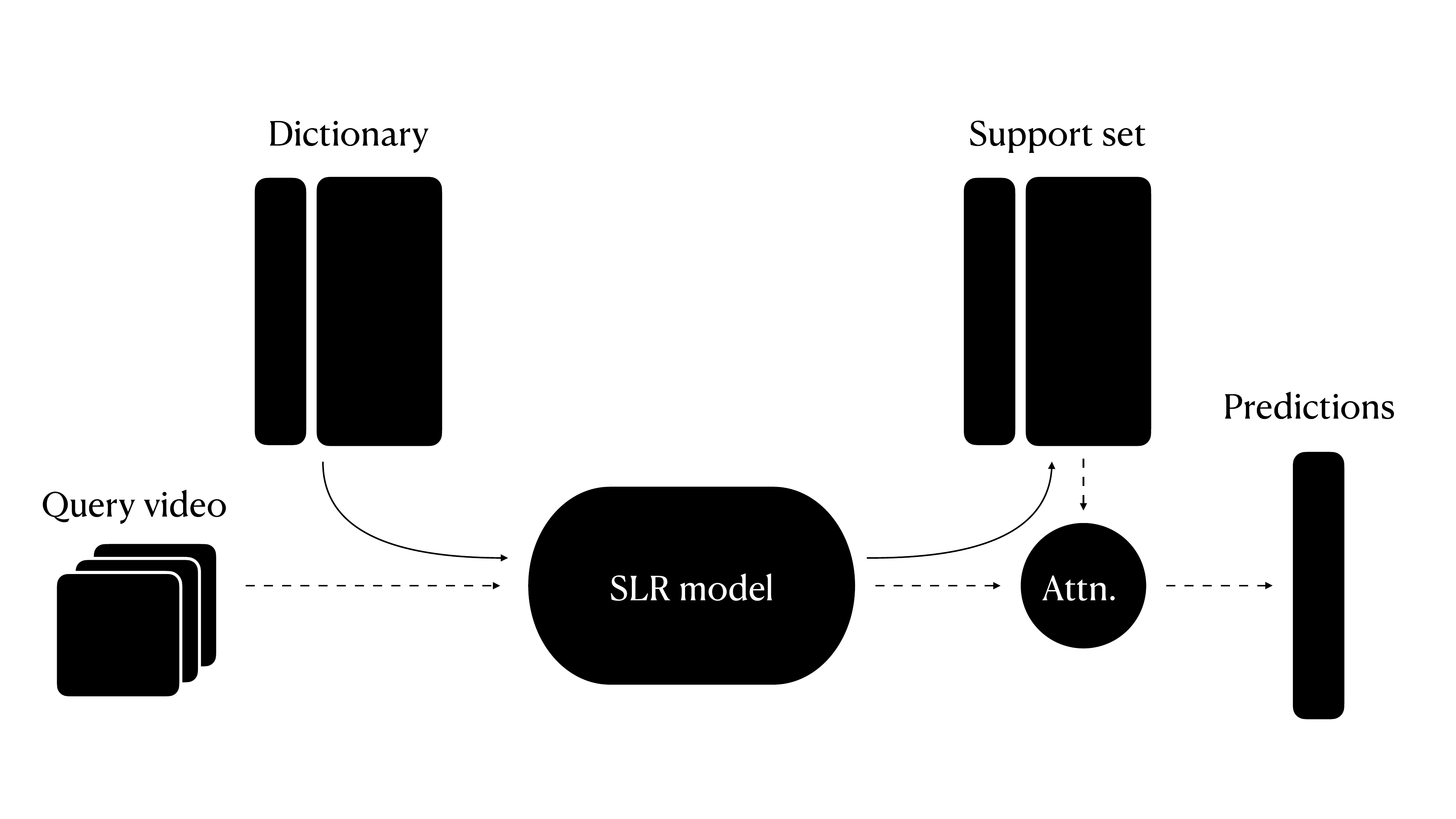}
    \caption{We perform one-shot sign classification to search through a dictionary with a query video.
    Solid arrows: a sign language dictionary is mapped to a support set of embeddings by an SLR model. This is done once.
    Dashed arrows: we can classify a new example (query the dictionary) by also mapping the example to an embedding with the same model and using attention to obtain probabilities for every label in the support set. This can be done without regenerating the support set.}
    \label{fig:matching}
\end{figure}

% \mathieu{je verwijst hier niet naar je figuur ;)}
% \mathieu{support set = exemplar set? beter om 1 en dezelfde term te gebruiken doorheen de paper dan}
Our one-shot inference exists out of two steps, namely: initialisation and inference. These steps are illustrated in \cref{fig:matching}. In the initialisation step -- illustrated with solid lines -- the frozen model converts all dictionary videos to embeddings. This collection of embeddings is known as the support set, which serves as a reference for subsequent comparisons during the inference phase. During the inference step -- depicted with dashed lines -- the model processes an input query video and generates its corresponding embedding. This embedding is then compared against the support set to determine the most similar entry. The comparison utilises the attention mechanism as described by \citep{bahdanau2014neural}, which effectively identifies the closest match from the support set solely based on the embeddings.

In essence, this method performs \emph{search} as described by \citet{mittal2021compositional}, who formulated it as the ``selection of a relevant entity from a set via \emph{query-key} interactions''. In our case, the queries and keys are the query videos and the support set respectively. Furthermore, this approach applies the softmax function, and hence the output can be interpreted as probabilities. These probabilities provide a measure of confidence in the match, which in turn allows for more nuanced decision-making, ultimately improving the system's interpretability and reliability.

We evaluate the one-shot classification method using two approaches. The first evaluation involves a limited query set with a large support set, assessing the model's ability to handle applications requiring extensive vocabularies, such as SLT. The second experiment introduces variability by randomly selecting different instances from a sign language dataset multiple times, effectively reconstructing the same classes within the support set. By measuring the spread of evaluation metrics across these variations, we investigate how the variability within sign categories and the selection of support set instances influence our approach.
% \mathieu{deze zin: ik zou verduidelijken wat je bedoelt met sign language training dataset, en dat het over de support set gaat. dus dat je een andere dictionary opstelt als het ware}

% \mathieu{goede segue naar datasets, maar ik vind de laatste zin vreemd}
Although this one-shot technique is remarkably elegant, the performance heavily relies on how the manual sign features are represented within the embedding. Two main factors independently impact this objective: the pretraining and the utilised datasets. We employ isolated sign language recognition for pretraining since it aligns best with our downstream goal. Furthermore, both the quality and quantity of the used pretraining data significantly impact the results of the downstream tasks, as the results in \cref{sec:oneshotresults} show.

\subsection{Datasets}
Several kinds of datasets are required to perform our experiments. First, we need one or more
pretraining datasets, on which we can train a model for ISLR.
Second, datasets for the one-shot ISLR evaluation are needed. These datasets include both real-world sign language dictionaries and existing isolated sign language datasets that we have adapted for the one-shot task.

\subsubsection{Pretraining Datasets} 
\label{sec:pretrainingdatasets}
% \mathieu{HIERDICTIONARY (zie comment ergens hierboven) - hier zou je kunnen zeggen dat dictionary search 1 vd applicaties is van one-shot, en dat je daarom met de2023querying vergelijkt}
One primary example of one-shot ISLR is dictionary retrieval, like the study by \cite{de2023querying}. To facilitate comparisons with this work, the same pretraining dataset is utilised. This dataset is derived from the VGT corpus
\citep{van2015het}. However, the limited number of classes and the severe class imbalance in this dataset may have contributed to suboptimal performance. Therefore, we look for a larger and more varied dataset for ISLR: we choose ASL Citizen \citep{desai2024asl} for its sizeable vocabulary and because DHH signers recorded the signs.

\begin{figure}
    \begin{subfigure}[b]{0.49\textwidth}
        \centering
        \includegraphics[width=\textwidth]{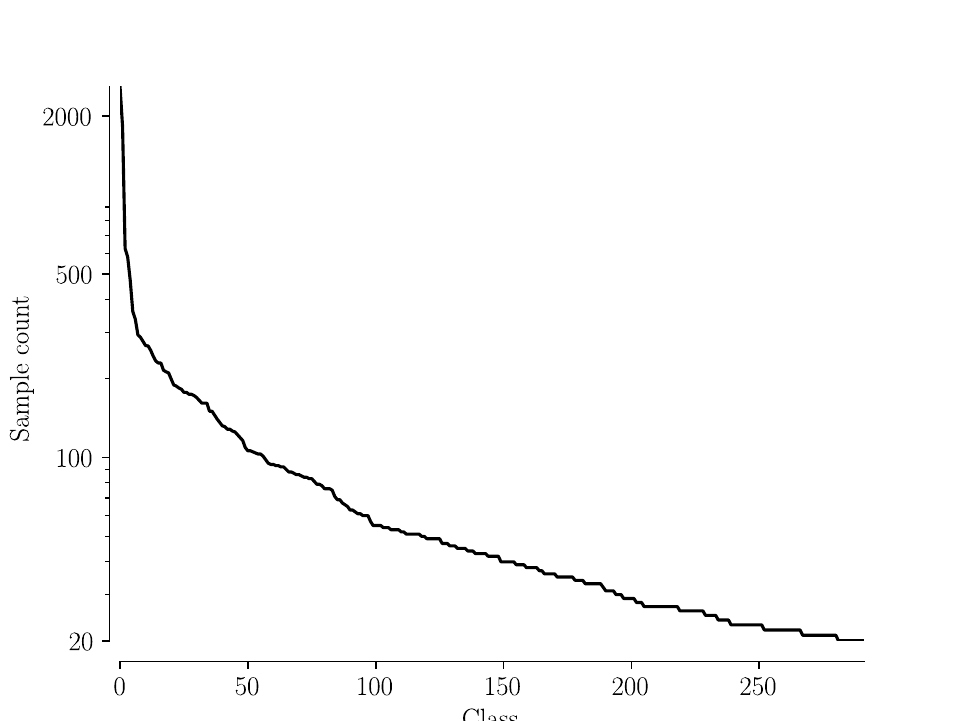}
        \caption{Corpus VGT \citep{van2015het}}
        \label{fig:distvgt}
    \end{subfigure}
    \hfill
    \begin{subfigure}[b]{0.49\textwidth}
        \centering
        \includegraphics[width=\textwidth]{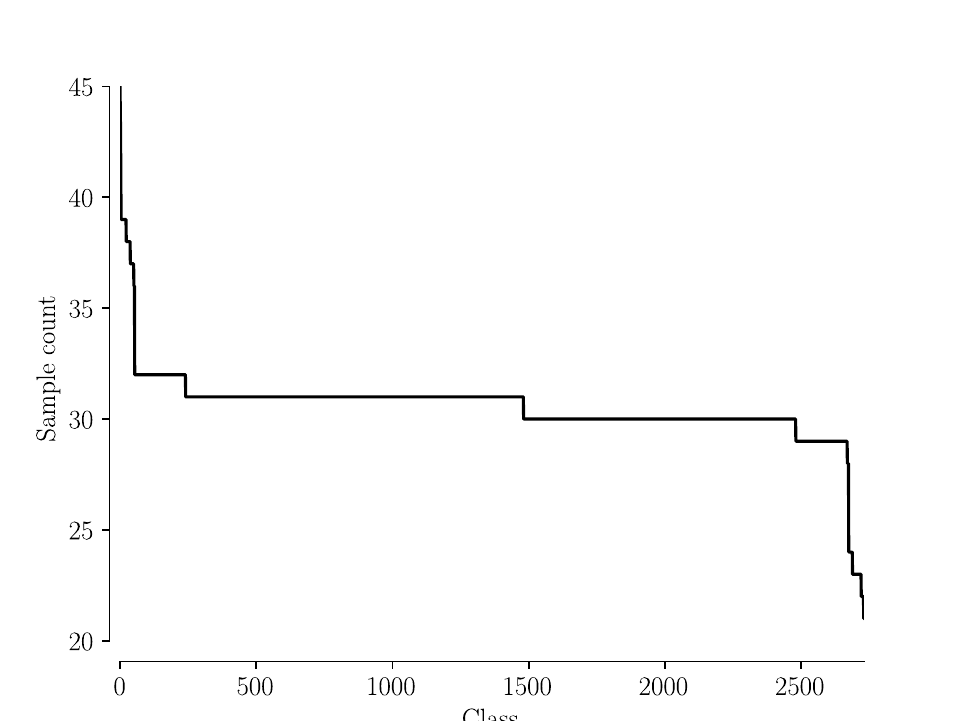}
        \caption{ASL-Citizen \citep{desai2024asl}}
        \label{fig:distasl}
    \end{subfigure}
    \caption{Class distributions of pretraining datasets, sorted by descending sample count.}
    \label{fig:distributions}
\end{figure}

The VGT dataset consists of 24,967 examples for 292 signs, but these examples are not uniformly
distributed (\cref{fig:distvgt}). Since the dataset is derived from a corpus that contains spontaneous language use,
the distribution is rather Zipfian. The majority of examples represent only a minority of the signs. The examples are cut from continuous signing, which means that every sign execution
is influenced by preceding and subsequent signs, a phenomenon referred to as co-articulation.

The ASL Citizen dataset is richer, containing 83,399 examples for 2,731 signs. These examples
are more uniformly distributed, with an average of 30 instances per sign (\cref{fig:distasl}).
Since the dataset entries were recorded in isolation, no co-articulation occurred. This more closely matches
the task of dictionary search. In fact, the ASL Citizen dataset was envisioned as a dataset of
``in-the-wild dictionary queries'' \citep{desai2024asl}.

% Both datasets are signer-independent, i.e., there is no overlap between the training, validation, and test sets in terms of the signers appearing in those sets. As such, data leakage is avoided and metric based evaluation gives a proper estimate of the performance of the models in the wild, when it is unlikely that someone from the training set will use the model.

%\toon{Misschien iets toevoegen over ASL-LEX, want dit haal ik later opnieuw aan in de limitaties.}
%\mathieu{vind ik hier niet per se nodig, maar doe maar als jij dit wel belangrijk vindt}

\subsubsection{Evaluation datasets} 
One application of our work is enabling native and non-native signers to search through sign language dictionaries by recording or uploading a video of a sign. To evaluate the performance of a 
% \mathieu{hier ook: mss zwijgen over dictionary search en enkel one shot? of expliciet vermelden dat het een toepassing is van one-shot}
dictionary search system (one-shot classifier) we always require two sets: dictionary entries, referred to as the support set, and dictionary queries, which serve as test examples. 
Both sets consist of short videos containing individual signs that mirror the entries
found in the dictionary. Ideally, these signs are performed in citation form, meaning that each sign starts and ends in a resting position. 

Due to sign language dialects and synonyms, multiple signs may correspond to a single word
or concept. However, we do not consider this as the same sign and instead label signs by unique IDs. In this work, we focus on retrieving the exact sign matching the query video, rather than its meaning in spoken language.

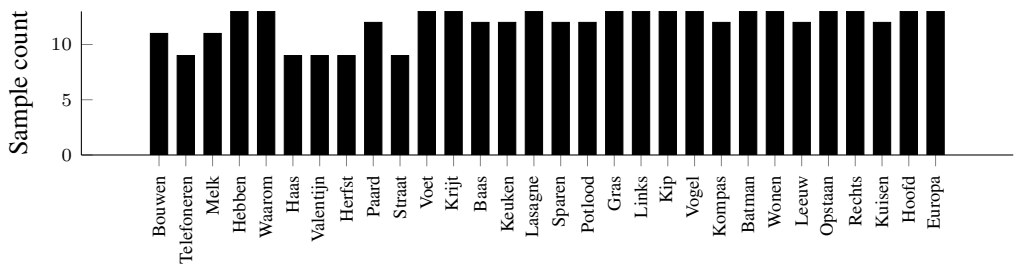
\begin{figure}
    \centering
    \begin{tikzpicture}
        \begin{axis}[
            ybar,
            width=\textwidth,
            height=0.25\textwidth,
            symbolic x coords={Bouwen, Telefoneren, Melk, Hebben, Waarom, Haas, Valentijn, Herfst, Paard, Straat, Voet, Krijt, Baas, Keuken, Lasagne, Sparen, Potlood, Gras, Links, Kip, Vogel, Kompas, Batman, Wonen, Leeuw, Opstaan, Rechts, Kuisen, Hoofd, Europa},
            xtick=data,
            nodes near coords={},
            ymin=0,
            ymax=13,
            axis y line*=left,
            axis x line*=bottom,
            xticklabel style={rotate=90, anchor=east},
            ylabel = {Sample count}
        ]
            % Add bars with hyperlinks
            \addplot [
                ybar,
                bar width=0.23cm,
                draw=black,
                fill=black,
                % enlarge x limits=0.2
                % xticklabel style={rotate=90, anchor=east}
            ] coordinates {(Bouwen, 11) (Telefoneren, 9) (Melk, 11) (Hebben, 13) (Waarom,13) (Haas,9) (Valentijn, 9) (Herfst,9) (Paard,12) (Straat, 9) (Voet, 13) (Krijt, 13) (Baas, 12) (Keuken, 12) (Lasagne, 13) (Sparen, 12) (Potlood, 12) (Gras, 13) (Links, 13) (Kip, 13) (Vogel, 13) (Kompas, 12) (Batman, 13) (Wonen, 13) (Leeuw, 12) (Opstaan,13) (Rechts, 13) (Kuisen, 12) (Hoofd, 13) (Europa,13)};

            \node [anchor=north] at (axis cs:Bouwen,0) {\href{\bouwen}{\rotatebox{90}{\phantom{Bouwen}}}};
            \node [anchor=north] at (axis cs:Telefoneren,0) {\href{\telefoneren}{\rotatebox{90}{\phantom{Telefoneren}}}};
            \node [anchor=north] at (axis cs:Melk,0) {\href{\melk}{\rotatebox{90}{\phantom{Melk}}}};
            \node [anchor=north] at (axis cs:Hebben,0) {\href{\hebben}{\rotatebox{90}{\phantom{Hebben}}}};
            \node [anchor=north] at (axis cs:Waarom,0) {\href{\waarom}{\rotatebox{90}{\phantom{Waarom}}}};
            \node [anchor=north] at (axis cs:Haas,0) {\href{\haas}{\rotatebox{90}{\phantom{Haas}}}};
            \node [anchor=north] at (axis cs:Valentijn,0) {\href{\valentijn}{\rotatebox{90}{\phantom{Valentijn}}}};
            \node [anchor=north] at (axis cs:Herfst,0) {\href{\herfst}{\rotatebox{90}{\phantom{Herfst}}}};
            \node [anchor=north] at (axis cs:Paard,0) {\href{\paard}{\rotatebox{90}{\phantom{Paard}}}};
            \node [anchor=north] at (axis cs:Straat,0) {\href{\straat}{\rotatebox{90}{\phantom{Straat}}}};
            \node [anchor=north] at (axis cs:Voet,0) {\href{\voet}{\rotatebox{90}{\phantom{Voet}}}};
            \node [anchor=north] at (axis cs:Krijt,0) {\href{\krijt}{\rotatebox{90}{\phantom{Krijt}}}};
            \node [anchor=north] at (axis cs:Baas,0) {\href{\baas}{\rotatebox{90}{\phantom{Baas}}}};
            \node [anchor=north] at (axis cs:Keuken,0) {\href{\keuken}{\rotatebox{90}{\phantom{Keuken}}}};
            \node [anchor=north] at (axis cs:Lasagne,0) {\href{\lasagne}{\rotatebox{90}{\phantom{Lasagne}}}};
            \node [anchor=north] at (axis cs:Sparen,0) {\href{\sparen}{\rotatebox{90}{\phantom{Sparen}}}};
            \node [anchor=north] at (axis cs:Potlood,0) {\href{\potlood}{\rotatebox{90}{\phantom{Potlood}}}};
            \node [anchor=north] at (axis cs:Gras,0) {\href{\gras}{\rotatebox{90}{\phantom{Gras}}}};
            \node [anchor=north] at (axis cs:Links,0) {\href{\links}{\rotatebox{90}{\phantom{Links}}}};
            \node [anchor=north] at (axis cs:Kip,0) {\href{\kip}{\rotatebox{90}{\phantom{Kip}}}};
            \node [anchor=north] at (axis cs:Vogel,0) {\href{\vogel}{\rotatebox{90}{\phantom{Vogel}}}};
            \node [anchor=north] at (axis cs:Kompas,0) {\href{\kompas}{\rotatebox{90}{\phantom{Kompas}}}};
            \node [anchor=north] at (axis cs:Batman,0) {\href{\batman}{\rotatebox{90}{\phantom{Batman}}}};
            \node [anchor=north] at (axis cs:Wonen,0) {\href{\wonen}{\rotatebox{90}{\phantom{Wonen}}}};
            \node [anchor=north] at (axis cs:Leeuw,0) {\href{\leeuw}{\rotatebox{90}{\phantom{Leeuw}}}};
            \node [anchor=north] at (axis cs:Opstaan,0) {\href{\opstaan}{\rotatebox{90}{\phantom{Opstaan}}}};
            \node [anchor=north] at (axis cs:Rechts,0) {\href{\rechts}{\rotatebox{90}{\phantom{Rechts}}}};
            \node [anchor=north] at (axis cs:Kuisen,0) {\href{\kuisen}{\rotatebox{90}{\phantom{Kuisen}}}};
            \node [anchor=north] at (axis cs:Hoofd,0) {\href{\hoofd}{\rotatebox{90}{\phantom{Hoofd}}}};
            \node [anchor=north] at (axis cs:Europa,0) {\href{\europa}{\rotatebox{90}{\phantom{Europa}}}};
        \end{axis}
    \end{tikzpicture}
    \caption{The number of dictionary queries per gloss is distributed approximately uniformly
    with mean 11.93. Each label on the horizontal axis is a link to the corresponding video within the Flemish Sign Language dictionary.}
    \label{fig:frequency}
\end{figure}

For the first evaluation set, we collect a set of dictionary queries in VGT for 30 distinct sign categories (\cref{fig:frequency}). The 10 categories used in prior dictionary retrieval research \citep{de2023querying} are a subset of our set of 30. As the first support set, we use a pre-existing sign language dictionary, that is, the VGT dictionary \citep{van2004woordenboek}.\footnote{The dictionary videos can be downloaded from this website: \url{https://taalmaterialen.ivdnt.org/download/woordenboek-vgt/}.}
This dictionary contains 10,235 unique signs at the time of writing. Like many other sign language dictionaries, every sign has exactly one example, which is why we opt for one-shot ISLR.

% Plot met count per gloss in evaluatie set
\newcommand{\bouwen}{https://woordenboek.vlaamsegebarentaal.be/gloss/BOUWEN?sid=1906}
\newcommand{\telefoneren}{https://woordenboek.vlaamsegebarentaal.be/gloss/TELEFONEREN?sid=11870}
\newcommand{\melk}{https://woordenboek.vlaamsegebarentaal.be/gloss/MELK?sid=7418}
\newcommand{\hebben}{https://woordenboek.vlaamsegebarentaal.be/gloss/HEBBEN?sid=4801}
\newcommand{\waarom}{https://woordenboek.vlaamsegebarentaal.be/gloss/WAAROM?sid=13564}
\newcommand{\haas}{https://woordenboek.vlaamsegebarentaal.be/gloss/HAAS?sid=16146}
\newcommand{\valentijn}{https://woordenboek.vlaamsegebarentaal.be/gloss/VALENTIJN?sid=16235}
\newcommand{\herfst}{https://woordenboek.vlaamsegebarentaal.be/gloss/HERFST?sid=4897}
\newcommand{\paard}{https://woordenboek.vlaamsegebarentaal.be/gloss/PAARD?sid=8880}
\newcommand{\straat}{https://woordenboek.vlaamsegebarentaal.be/gloss/STRAAT?sid=11560}
\newcommand{\voet}{https://woordenboek.vlaamsegebarentaal.be/gloss/VOET?sid=13229}
\newcommand{\krijt}{https://woordenboek.vlaamsegebarentaal.be/gloss/KRIJT?sid=6398}
\newcommand{\baas}{https://woordenboek.vlaamsegebarentaal.be/gloss/BAAS?sid=827}
\newcommand{\keuken}{https://woordenboek.vlaamsegebarentaal.be/gloss/KEUKEN?sid=5851}
\newcommand{\lasagne}{https://woordenboek.vlaamsegebarentaal.be/gloss/LASAGNE?sid=6637}
\newcommand{\sparen}{https://woordenboek.vlaamsegebarentaal.be/gloss/SPAREN?sid=11147}
\newcommand{\potlood}{https://woordenboek.vlaamsegebarentaal.be/gloss/POTLOOD?sid=9508}
\newcommand{\gras}{https://woordenboek.vlaamsegebarentaal.be/gloss/GRAS?sid=4490}
\newcommand{\links}{https://woordenboek.vlaamsegebarentaal.be/gloss/LINKS?sid=6911}
\newcommand{\kip}{https://woordenboek.vlaamsegebarentaal.be/gloss/KIP?sid=5932}
\newcommand{\vogel}{https://woordenboek.vlaamsegebarentaal.be/gloss/VOGEL?sid=13258}
\newcommand{\kompas}{https://woordenboek.vlaamsegebarentaal.be/gloss/KOMPAS?sid=6212}
\newcommand{\batman}{https://woordenboek.vlaamsegebarentaal.be/gloss/BATMAN?sid=15821}
\newcommand{\wonen}{https://woordenboek.vlaamsegebarentaal.be/gloss/WONEN?sid=14127}
\newcommand{\leeuw}{https://woordenboek.vlaamsegebarentaal.be/gloss/LEEUW?sid=6690}
\newcommand{\opstaan}{https://woordenboek.vlaamsegebarentaal.be/gloss/OPSTAAN?sid=8682}
\newcommand{\rechts}{https://woordenboek.vlaamsegebarentaal.be/gloss/RECHTS?sid=9803}
\newcommand{\kuisen}{https://woordenboek.vlaamsegebarentaal.be/gloss/KUISEN?sid=6464}
\newcommand{\hoofd}{https://woordenboek.vlaamsegebarentaal.be/gloss/HOOFD?sid=5069}
\newcommand{\europa}{https://woordenboek.vlaamsegebarentaal.be/gloss/EUROPA?sid=3646}

The second set of evaluation datasets includes AUTSL \citep{sincan2020autsl} and WLASL \citep{li2020word}, which consist of Turkish and American Sign Language respectively. The WLASL dataset provides splits for various dictionary sizes -- 100, 300, 1000, and 2000 signs -- offering a robust evaluation for scaling dictionary sizes. AUTSL, on the other hand, comprises a vocabulary of 226 independent signs. For all experiments using these datasets, the test set serves as queries, while the support set is constructed by randomly selecting one entry per class from the training set for 100 different times. To ensure these sets were different, seeds were used. Random sampling provides a more rigorous assessment of the model's performance, demonstrating that its performance does not depend on the quality or consistency of individual dictionary entries.

% \mathieu{@Toon, jij hebt de URLs van al deze videos in je broncode staan vermoed ik? Kunnen we deze in appendix toevoegen aan de paper? Zoals in mijn PhD, gewoon per gloss de link naar de video in het WDB}

% \mathieu{@Toon, kan je hier in een tabel de statistieken van hoeveel voorbeelden we hebben per gebaar oplijsten? aangezien we toch zoveel plaats hebben, is 1 key frame per gebaar in een figuur
% misschien ook wel tof, daar kan ik evt voor zorgen}

% \begin{figure}[ht!]
%     \centering
%     % \includesvg[width=\textwidth]{fig/frequency2_0.svg}
%     \includegraphics[width=\textwidth]{fig/frequency2_0_axes_1.pdf}
%     \caption{The number of dictionary queries per gloss is distributed approximately uniformly
%     with mean 11.93.}
%     \label{fig:frequency}
% \end{figure}

\subsection{Pretraining: Keypoint-based Sign Language Recognition}
\label{sec:pretraining}

To perform one-shot sign language recognition, we first need to pretrain a sign language recognition model. We choose to use a keypoint-based model. By performing human keypoint
estimation on the sign language videos, the sign language recognition task is facilitated.
Moreover, we hypothesise that using a keypoint-based model is beneficial to the one-shot
classification task, because it reduces the impact of the visual properties (background, lighting,
clothing, etc.) of the query and the dictionary videos, and it better aligns the input
distributions of the data used in the pretraining step and the querying step. Another advantage of using keypoints is the ability to integrate publicly available estimators, such as MediaPipe, into client-side applications, ensuring both privacy and low-latency communication.

More specifically, we choose to employ MediaPipe \citep{grishchenko2020mediapipe} for the reasons listed in \cref{sec:islr}. MediaPipe predicts the pose, hands and face keypoints. We only utilise the pose and hand information. Thus, our approach solely focuses on the manual components of signs: handshape, movement, place of articulation, and orientation. These manual features are transferable across sign languages. 
% Mouthings are more sign language specific \citep{bank2015alignment}; we do not integrate them in this work. 
% Models processing the mouthings were trained, optimized and tested, but lowered our one-shot results. 
We also trained, optimised and tested embedding models that use mouth keypoints, but the results for one-shot recovery were worse.
This decrease can be attributed to the fact that mouthings are more language-specific \citep{bank2015alignment}. Therefore, we do not deem them essential to this work, but acknowledge their importance in broader sign language processing.

% We choose the ISLR model from the SignON research project \citep{holmes2023scarcity}, because
% training and evaluation code are available under the Apache 2.0 licence.\footnote{\url{https://github.com/signon-project/wp3-slr-pipeline} and \url{https://github.com/m-decoster/VGT-SL-Dictionary}} The weights
% are also available online.\footnote{\url{https://huggingface.co/signon-project/slr-poseformer-vgt}}
% The name of this model is ``PoseFormer''.
% Previous research with similar models has also illustrated that their constituent sub-networks
% can be transferred to different languages without finetuning \citep{de2023towards}, which is
% a useful property for our use case.

\begin{figure}[ht!]
    \centering
    \includegraphics[clip,trim=2pt 10pt 10pt 10pt,width=\textwidth]{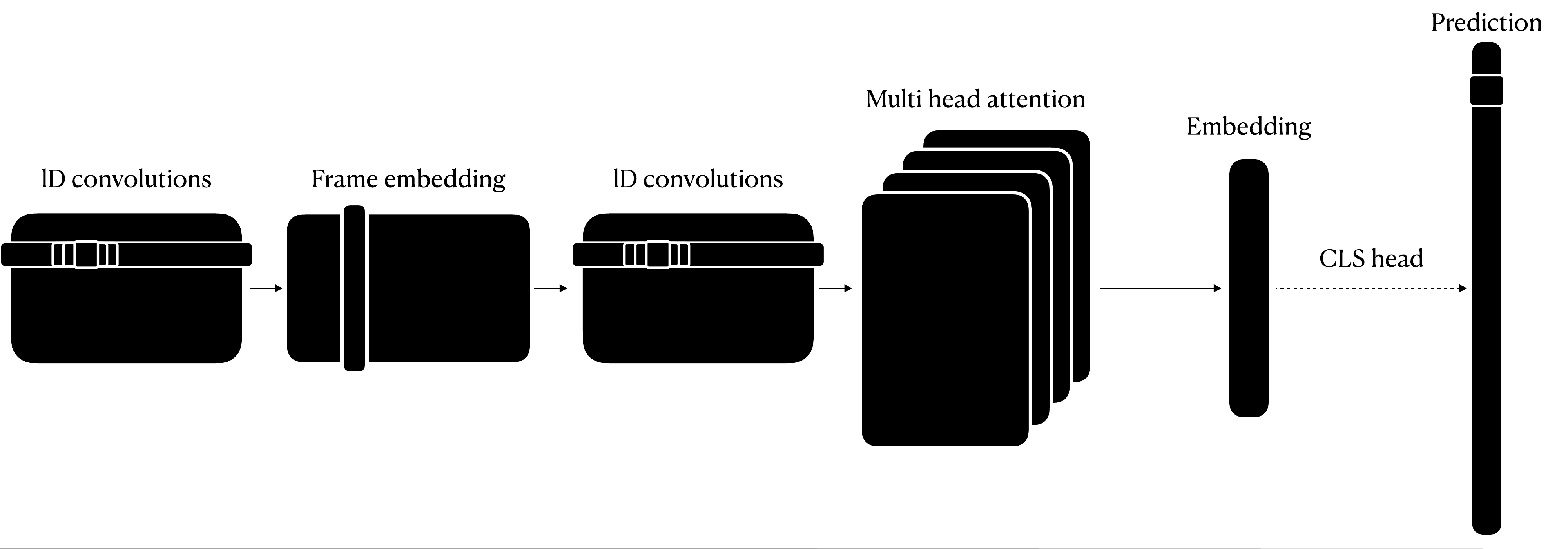}
    \caption{The PoseFormer model, represented by solid lines, consists of several blocks. The 1D convolutions process data along the temporal axis, while the frame embedding block handles individual frames. Finally, the multi-head attention block extracts relevant features. After training, the classification head, consisting of a single linear layer (depicted with a dashed line), is removed.}
    \label{fig:pf}
\end{figure}

We build on the architectural style of the SignON research project \citep{holmes2023scarcity}, making slight modifications to the number of layers and blocks to better fit the requirements of our task. A schematic overview of this so-called PoseFormer network is shown in \cref{fig:pf}. The network integrates dense and convolutional blocks, followed by a multi-head attention mechanism.

In this model, a sequence of keypoint coordinates representing human poses serves as input. Initially, 1D convolutions are used for temporal smoothing, capturing short-term dependencies in the sequence. The output is then fed into a multi-layer dense sub-network, which processes each frame individually to extract non-linear representations of the pose. These frame-level representations are subsequently passed through a convolutional block that focuses on learning local temporal context. Finally, a multi-head self-attention mechanism captures global temporal dependencies across the sequence. The resulting vector, representing each sign, is input into a linear classifier for final prediction.

\begin{table}[ht!]
    \caption{We optimise hyperparameters for both models}
    \label{tab:hyperparams}
    \centering
    \begin{tabular}{l|ll}
         \hline
         \textbf{Hyperparameter} & \textbf{ASL} & \textbf{VGT} \\
         \hline
         Batch size & 64 & 128 \\
         Learning rate & 0.0003 & 0.0003 \\
         Representation size & 160 & 192 \\
         Attention layers & 4 & 4 \\
         Attention heads & 8 & 8 \\
         Dropout & 0.2 & 0.2 \\
         \hline
    \end{tabular}
\end{table}

As noted in \cref{sec:pretrainingdatasets}, we use two pretraining datasets with different properties and compare results obtained with them. When we pretrain the PoseFormer with the VGT dataset, we refer to this model as \PFVGT, and when we pretrain it with the ASL Citizen dataset, we refer to it as \PFASL. 
% We optimise hyperparameters for the \PFASL. 
% For the \PFVGT, since we use the pretrained checkpoint from the SignON project, we do not optimise hyperparameters. 
The used hyperparameters for both datasets are represented in \cref{tab:hyperparams}.

% It is a well-known best practice to use separate training, validation, and test sets in deep learning. To report the pretraining stage, we adhere to the predefined splits of the provided dataset. However, for our dictionary search task, we do not rely on the ASL Citizen test set for evaluation. Instead, we add it to our training set to maximally leverage the available data, nearly doubling the size of the training set, from 40,154 to 73,095.

\subsection{Evaluation Metrics}
Several evaluation metrics are used throughout the entire pipeline to ensure a robust evaluation of the model. First, we report the mean Recall@$K$ for $K \in [1, 5, 10]$. Furthermore, two ranking metrics are employed: mean reciprocal rank (MRR) and normalised discounted cumulative gain (nDCG). Given that there is only one relevant item per prediction (its label), the interpretations of MRR and nDCG are similar in this scenario. The key difference is that the inverse of the MRR is the harmonic mean of the ranks of all predictions, which provides an alternative view of the results. For all three metrics, higher is better and a value of one indicates optimal performance. All three metrics -- Recall@$K$, MRR, and nDCG -- are used to evaluate the pretraining, to allow for comparison with a baseline ISLR model. For the evaluation of the one-shot methods, reporting only the Recall@$K$ and the MRR is deemed sufficient, since the interpretation of MRR and nDCG is very similar.

% Let $r_i$ be the rank for the $i$\textsuperscript{th} test example, that is, the one-based index of the ground truth label in the ordered list of model predictions. Then, for a set of $N$ test examples,
% \begin{equation}
%     \mathrm{MRR} = \frac{1}{N}\sum_{i=1}^{N}\frac{1}{r_i}.
% \end{equation}

% nDCG is similar to MRR, but it considers the ranks of all correct predictions, favouring relevant results that occur earlier in the ordered list of model predictions. Since in our case search results are either correct or incorrect,
% we define the relevance of a search result as a binary value: it is $1$ when the prediction $\hat{y}$
% is equal to the ground truth label $y$, and $0$ otherwise.
% The DCG can be computed for a single test example by considering the ordered list of $M$ predictions.
% For an expected ground truth label $y$, the DCG is equal to
% \begin{equation}
%     \mathrm{DCG} = \sum_{j=1}^{M}\frac{\mathbbm{1}_{\hat{y}_j = y}}{\log_2(j+1)}.
%     \label{eq:dcg}
% \end{equation}
% nDCG is the normalised version of DCG. This normalisation is done by dividing the DCG by the ideal DCG (IDCG). IDCG is the DCG of the optimal ranking. In our specific case, with only one relevant item, the IDCG is always equal to one. Therefore the nDCG simplifies directly to DCG:
% \begin{equation}
%     \mathrm{nDCG} = \frac{\mathrm{DCG}}{\mathrm{IDCG} } = \mathrm{DCG}.
%     \label{eq:ndcg}
% \end{equation}
% We compute the DCG for all $N$ test examples and report the mean.

\section{Results}
\label{sec:results}

\subsection{Pretraining}

The pretraining stage results are summarised in table \ref{tab:asl_results}. The I3D model \citep{desai2024asl} is considered the baseline: it is, as of writing, the best-performing model in the scientific literature for ASL Citizen.
We compare the PoseFormer model with this baseline. 
% Since this model has additional components compared to the Poseformer Transformer Network (PTN) used by \citet{de2023querying} for one-shot ISLR, 
% We perform a limited ablation study to determine the importance of the individual components of the network.
% \mathieu{vind je deze ablation nog nuttig? indien plaats tekort vind ik dit een kandidaat om te schrappen}
To further assess the PoseFormer’s architecture, we conduct a limited ablation study, isolating the impact of its key components. 
Specifically, these components are: the input convolutions and intermediate convolutions, which
appear respectively before and after the frame embedding sub-network. We remove these components
individually and measure the impact with respect to our metrics.
% \mathieu{@Toon: wat mij hier niet duidelijk is is, drop je telkens 1 deel of drop je er meerdere (cumulatief)? Ik vind ook die embedding size discussie hier niet super relevant, ik zou gewoon de beste houden}
%\toon{Enkel poseformer 160 behouden, maar die voor 192 staan ook nog in latex moesten we ze nog nodig hebben. Je weet maar nooit.}

\begin{table}[ht!]
    \caption{The PoseFormer outperforms the I3D baseline on the pretraining task (ASL Citizen). The ablation study
    illustrates the importance of the frame embedding and convolutions in the PoseFormer.}
    \label{tab:asl_results}
    \centering
    \begin{tabularx}{\textwidth}{l | X X | X X X}
         \hline
         \textbf{Model} & $\uparrow$ \textbf{MRR} & $\uparrow$ \textbf{nDCG} & $\uparrow$ \textbf{Rec{@}1} & $\uparrow$ \textbf{Rec{@}5} & $\uparrow$ \textbf{Rec{@}10}\\
         \hline
         % Poseformer 192: & 0.8287 & 0.8672 & 0.7456 & 0.9287 & 0.9531\\
         % \hspace{4mm}Input convolution & 0.8229 & 0.8624 & 0.7401 & 0.9239 & 0.9489 \\
         % \hspace{4mm}Frame embedding & 0.8151 & 0.8564 & 0.7289 & 0.9194 & 0.9470 \\
         % \hspace{4mm}Intermediate convolution & 0.8222 & 0.8621 & 0.7362 & 0.9271 & 0.9514 \\
         Poseformer & \textbf{0.833} & \textbf{0.870} & \textbf{0.751} & \textbf{0.932} & 0.955 \\
         \hspace{2mm}No input conv. & 0.831 & 0.869 & 0.749 & 0.931 & \textbf{0.956} \\
         \hspace{2mm}No frame embedding & 0.811 & 0.854 & 0.723 & 0.920 & 0.946 \\
         \hspace{2mm}No intermediate conv. & 0.819 & 0.860 & 0.733 & 0.926 & 0.951 \\
         \hline
         I3D \citep{desai2024asl} & 0.733 & 0.791 & 0.631 & 0.861 & 0.909\\
         \hline
         % \hline
    \end{tabularx}
\end{table}

% \begin{table}[ht!]
%     \caption{The PoseFormer outperforms the I3D baseline on the pretraining task (ASL Citizen). The ablation study
%     illustrates the importance of the frame embedding and convolutions in the PoseFormer.}
%     \label{tab:asl_results}
%     \centering
%     \begin{tabularx}{\textwidth}{l | X X | X X X}
%          \hline
%          \textbf{Model} & $\uparrow$ \textbf{MRR} & $\uparrow$ \textbf{DCG} & $\uparrow$ \textbf{Rec{@}1} & $\uparrow$ \textbf{Rec{@}5} & $\uparrow$ \textbf{Rec{@}10}\\
%          \hline
%          % Poseformer 192: & 0.8287 & 0.8672 & 0.7456 & 0.9287 & 0.9531\\
%          % \hspace{4mm}Input convolution & 0.8229 & 0.8624 & 0.7401 & 0.9239 & 0.9489 \\
%          % \hspace{4mm}Frame embedding & 0.8151 & 0.8564 & 0.7289 & 0.9194 & 0.9470 \\
%          % \hspace{4mm}Intermediate convolution & 0.8222 & 0.8621 & 0.7362 & 0.9271 & 0.9514 \\
%          Poseformer & \textbf{0.8325} & \textbf{0.8702} & \textbf{0.7509} & \textbf{0.9319} & 0.9552 \\
%          \hspace{2mm}No input conv. & 0.8311 & 0.8691 & 0.7492 & 0.9306 & \textbf{0.9555} \\
%          \hspace{2mm}No frame embedding & 0.8114 & 0.8535 & 0.7231 & 0.9196 & 0.9461 \\
%          \hspace{2mm}No intermediate conv. & 0.8193 & 0.8599 & 0.7329 & 0.9260 & 0.9507 \\
%          \hline
%          Baseline: I3D \cite{desai2024asl} & 0.7332 & 0.7913 & 0.6310 & 0.8609 & 0.9086\\
%          \hline
%          % \hline
%     \end{tabularx}
% \end{table}

\Cref{tab:asl_results} illustrates the PoseFormer's significant improvement over the I3D baseline \citep{desai2024asl} (+ 0.0993 MRR and + 0.1199 Recall@1). As of writing, our results are the state of
the art on the ASL Citizen dataset.
The ablation study emphasises the importance of the components of the PoseFormer.
The input convolutions, which act as temporal filters on the raw keypoint coordinate features, have limited impact on the scores. The frame embedding is also present in the original network, and has a
larger impact (+ 0.0211 MRR and + 0.0278 Recall@1). The intermediate convolutions, which appear in the network between
the frame embedding and self-attention blocks, also have a larger impact than the input convolutions
(+ 0.0132 MRR and + 0.018 Recall@1). These results confirm the relevance of the individual components of the PoseFormer Network.

% The intermediate convolutions have a larger impact (
% Furthermore, each component plays a demonstrable role. The intermediate convolution and frame embedding significantly impact performance. The frame embedding captures crucial pose information for accurate sign recognition. The intermediate convolution smoothes these representations, ultimately mitigating poor landmark representations. The input convolution layer seems less critical. This input filtering might be more relevant for noisier datasets, like corpus-VGT \cite{van2015het}. However, its inclusion allows for a fair comparison to the results of the model used by De Coster and Dambre \cite{de2023querying} in the following section.

\subsection{One-Shot Sign Language Recognition} 
\label{sec:oneshotresults}

There are two sets of results for the one-shot ISLR. The first considers the \PFASL model's capabilities to handle extremely large dictionaries. The second set evaluates both models' resilience to the selection of support set samples. For this set of experiments, we only consider the Recall@$K$ and the MRR, since the interpretation of MRR and nDCG is very similar. 
% \mathieu{dit vond ik niet zo heel duidelijk uit het methodology gedeelte, kan je dat daar nog duidelijker maken wat er precies gebeurt? ik heb daar ook een opmerking gezet denk ik}

\subsubsection{Large dictionaries}

\begin{figure}[ht!]
    \begin{subfigure}[b]{0.49\textwidth}
        \centering
        \includegraphics[width=\textwidth]{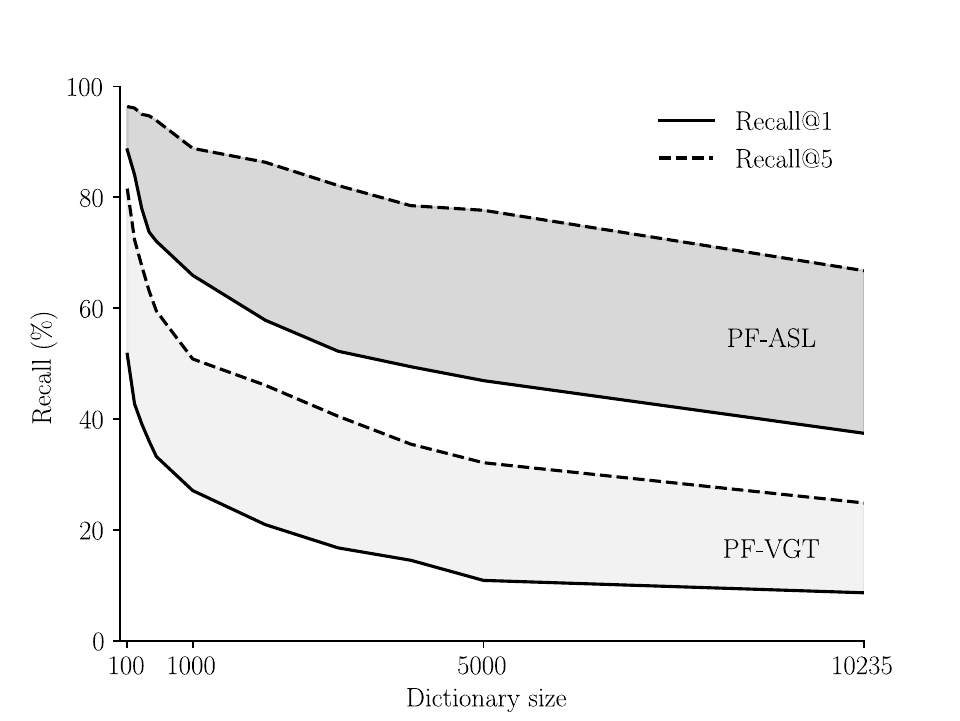}
        \caption{Recall@1 and Recall@5 for large dictionary search}
        \label{fig:top_1_vs_top_5}
    \end{subfigure}
    \hfill
    \begin{subfigure}[b]{0.49\textwidth}
        \centering
        \includegraphics[width=\textwidth]{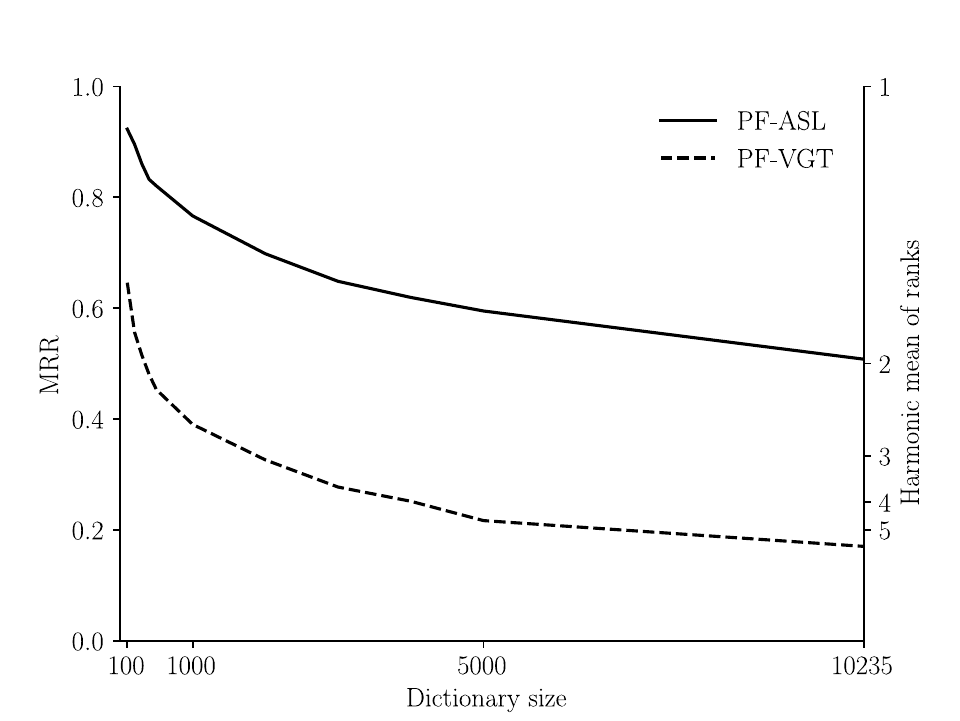}
        \caption{MRR for large dictionary search}
        \label{fig:mrr}
    \end{subfigure}
    \caption{Metrics for one-shot evaluation}
    \label{fig:ranking_metrics}
\end{figure}

\Cref{fig:top_1_vs_top_5} displays the relation between the dictionary size and Recall@$K$. We refer to the results of the independent models using the names from \cref{sec:pretraining}: \PFVGT~and \PFASL~(i.e., the PoseFormer pretrained on VGT and ASL respectively). For a dictionary of 100 signs, \PFASL~has a Recall@1 of 0.888. \PFVGT, despite being pretrained on the language of this dictionary lookup task, achieves a score of only 0.506.
% \mathieu{ik heb de volgende zin uitgedaan, want ik vind het wat appels met peren vergelijken. Ik snap wat je ermee wil aantonen. Ik zou het dan eerder bij MRR bespreken, waar je kan zeggen dat
% MRR ASL @ 10235 > MRR VGT @ 100. dan heb je het over dezelfde metriek en niet Recall@1 bij het ene model en Recall@2 bij het andere}
% To put this into perspective, the proposed method scores 0.5028 top-2 recall for dictionary size 10235.
For the full dictionary of 10,235 signs, \PFASL~achieves a Recall@1 score of 0.374, which is a substantial improvement over the 0.089 score obtained by \PFVGT. To put these results into perspective, a random search through the 10,235 classes would yield a Recall@1 of only 9.77e-5. These results underscore the impressive performance of \PFASL~in accurately predicting the correct class from a vast dictionary.

For the \PFASL, although not optimal, the MRR (\cref{fig:mrr}) remains relatively high at 0.508 across 10,235 categories. This suggests that, on average, the correct prediction ranks second. On the other hand, Recall@5 offers a complementary perspective: for the same 10,235 categories, the correct result is among the top five predictions 67\% of the time. This demonstrates the model's ability to consistently rank the correct prediction within a small set of top choices.

% The MRR is shown in \cref{fig:mrr}. For the \PFASL, it is not perfect, but high (still 0.508 for 10,235 categories). This illustrates that on average, the correct prediction appears at rank 2. The Recall@5 gives a different interpretation of this: for 10,235 categories, the correct result is in the top 5 predictions 67\% of the time. 

% \begin{figure}[ht!]
%         \centering
%         \includesvg[width=0.7\textwidth]{fig/compare_acc.svg}
%         \caption{Recall@1 and Recall@5 for large dictionary search}
%         \label{fig:top_1_vs_top_5}
% \end{figure}

% \begin{figure}
%     \begin{subfigure}[b]{0.49\textwidth}
%         \centering
%         \includesvg[width=\textwidth]{fig/top_1_acc.svg}
%         \caption{Recall@1 for dictionary search}
%         \label{fig:top_1}
%     \end{subfigure}
%     \hfill
%     \begin{subfigure}[b]{0.49\textwidth}
%         \centering
%         \includesvg[width=\textwidth]{fig/top_5_acc.svg}
%         \caption{Recall@5 for dictionary search}
%         \label{fig:top_5}
%     \end{subfigure}
%     \caption{Recall@$K$ accuracy for dictionary search}
%     \label{fig:dict_search}
% \end{figure}

% \mathieu{TODO: EXACTE NUMMERS 0.5 en 80\%}

% \toon{top 1-10 op 10235: [37.4301676 , 50.27932961, 58.10055866, 64.24581006, 66.75977654,69.55307263, 72.34636872, 74.58100559, 77.37430168, 78.49162011]}

\subsubsection{Support set perturbation}

The results on the support set perturbation are given in table \ref{tab:pertubation_results}. We only report the Recall@1 and MRR for both the \PFVGT~and \PFASL. Once again, these results highlight that the \PFASL~is a robuster model than the \PFVGT, because the pretraining dataset better aligns with the downstream task and contains more different signs.
% \mathieu{"despite..." -> in plaats daarvan zegggen dat het door de pretraining dataset komt "because the pretraining dataset better aligns with the downstream task and is more general" of zo}
On the AUTSL dataset, we achieved a Recall@1 of 58.1\%, while the \PFVGT~only achieved a score as high as 40.1\%. Although the results differ by almost 20\%, the spread on both results is about 3\%. The same can be observed in the MRR.
% \mathieu{waar ik het zag heb ik kommagetallen van bv 58,1 aangepast naar 58.1 maar een final pass zou nog goed kunnen zijn}

\begin{table}[ht!]
    \caption{To account for the evolving nature of sign languages, the PoseFormers were tested across various languages. The table also presents the standard deviation associated with perturbations in the dictionary.}
    \label{tab:pertubation_results}
    \centering
    % \scriptsize
    \begin{tabularx}{\textwidth}{l | X X | X X}
         \hline
         % &&&&\\[-1.8mm]
         & \multicolumn{2}{c|}{\PFASL} & \multicolumn{2}{c}{\PFVGT}  \\   
         \hline
         % &&&&\\[-1.8mm]
         \textbf{Dataset} & $\uparrow$ \textbf{Recall@1} & $\uparrow$ \textbf{MRR} & $\uparrow$ \textbf{Recall@1} & $\uparrow$ \textbf{MRR} \\
         \hline 
         % &&&&\\[-2mm]
         WLASL100 & 0.614 $\pm$ 0.024 & 0.692 $\pm$ 0.020 & 0.384 $\pm$ 0.017 & 0.485 $\pm$ 0.017 \\
         WLASL300 & 0.558 $\pm$ 0.011 & 0.646 $\pm$ 0.010 & 0.298 $\pm$ 0.008 & 0.397 $\pm$ 0.009 \\
         WLASL1000 & 0.470 $\pm$ 0.006 & 0.569 $\pm$ 0.005 & 0.225 $\pm$ 0.003 & 0.308 $\pm$ 0.004 \\
         WLASL2000 & 0.391 $\pm$ 0.004 & 0.500 $\pm$ 0.004 & 0.198 $\pm$ 0.002 & 0.269 $\pm$ 0.003 \\
         \hline
         AUTSL & 0.581 $\pm$ 0.036 & 0.687 $\pm$ 0.032 & 0.401 $\pm$ 0.034 & 0.533 $\pm$ 0.033 \\
         \hline
         %    % WLASL &&&&\\
         %    WLASL100 & 0.614 $\pm$ 0.024 & 0.692 $\pm$ 0.020 & 0.755 $\pm$ 0.016 & 0.384 $\pm$ 0.017 & 0.485 $\pm$ 0.017  & 0.586 $\pm$ 0.014 \\
         %    WLASL300 & 0.558 $\pm$ 0.011 & 0.646 $\pm$ 0.010 & 0.715 $\pm$ 0.009 & 0.298 $\pm$ 0.008 & 0.397 $\pm$ 0.009  & 0.508 $\pm$ 0.008 \\
         %    WLASL1000 & 0.470 $\pm$ 0.006 & 0.569 $\pm$ 0.005 & 0.651 $\pm$ 0.004 & 0.225 $\pm$ 0.003 & 0.308 $\pm$ 0.004  & 0.425 $\pm$ 0.003 \\
         %    WLASL2000 & 0.391 $\pm$ 0.004 & 0.500 $\pm$ 0.004 & 0.593 $\pm$ 0.003 & 0.198 $\pm$ 0.002 & 0.269 $\pm$ 0.003  & 0.386 $\pm$ 0.002 \\
         % \hline
         % &&&&\\[-1.8mm]
         %    AUTSL & 0.581 $\pm$ 0.036 & 0.687 $\pm$ 0.032 & 0.755 $\pm$ 0.026 & 0.401 $\pm$ 0.034 & 0.533 $\pm$ 0.033 & 0.633 $\pm$ 0.027 \\
         % \hline
    \end{tabularx}
\end{table}

Secondly, we evaluated the WLASL dataset across all predefined dictionary sizes. As expected, test performance diminished with increasing dictionary sizes, with Recall@1 dropping from 0.611 to 0.391. On the largest dictionary size, we also achieved an MRR of 0.500. This means that, on average, the correct prediction ranks second. In comparison, \PFVGT~places the correct prediction at rank 4.

The results also show a consistently small standard deviation across both datasets, models, and all dictionary sizes, indicating stable performance regardless of dataset variability. This consistency suggests that one-shot ISLR classification is robust to differences in data, maintaining reliable performance across various scenarios. Interestingly, the standard deviation decreases as dictionary size increases, likely due to the larger evaluation set offering a more comprehensive assessment of model accuracy. These findings reinforce the robustness of the models and their strong generalisation capabilities, making them well-suited for real-world applications across diverse settings.

% \toon{vind dit misschien een abrupt einde, weet niet wat nog toe te voegen. Misschien nog iets zeggen over de stds?}\mathieu{ja, want dat is eigenlijk het punt dat je wilt maken: de spread is zeer klein dus is het model robust voor dictionary entry selection}

\section{Discussion}

The results reveal four key insights. First, we confirm that keypoint-based models can achieve state-of-the-art results on the challenging isolated sign language recognition task. By utilising keypoints, the one-shot classification task becomes more feasible, as the pose estimator eliminates visual differences between datasets. Second, we demonstrate that the size, vocabulary, and class distribution of the dataset are critical for the pretraining phase, significantly influencing downstream performance in one-shot classification. Next, we find that alignment between the pretraining and downstream languages is less important than these dataset characteristics. Finally, our results highlight that representing individual signs, rather than relying on translations, is feasible and also crucial for creating future-proof sign language technologies. We detail these insights below.

%\subsubsection{Keypoint-Based Models Achieve State-of-the-art Performance}
Indeed, the keypoint-based PoseFormer achieves state-of-the-art results on the challenging large-vocabulary ASL Citizen dataset. It outperforms the I3D baseline by 0.120 Recall@1 (a 19\% increase). Moreover, the model also transfers seamlessly to downstream tasks on different languages, such as vector-based dictionary search for VGT, as our results illustrate.

%\subsubsection{The Pretraining Step is Crucial for the One-Shot Task}
The model's performance in the one-shot setting depends on the richness of the pretraining data.
Two specific traits of the ASL Citizen dataset enable the high performance of our one-shot
classification approach.  First, it encompasses a broad vocabulary of 2,731 unique glosses.
% , compared to just 250 glosses in other datasets like ASL Signs \cite{asl-signs}\mathieu{misschien wat raar om dit te vermelden aangezien we verder niks doen met ASL Signs. Gewoon zeggen: broad vocabulary punt}. Despite having a similar number of examples -- 83,399 for ASL Citizen versus 94,477 for ASL Signs -- the larger vocabulary in ASL Citizen results in fewer samples per class. 
This large and varied gloss set ensures exposure to a diverse range of signs, which is critical for robust model performance.
Second, the ASL Citizen dataset maintains a uniform class distribution, offering sufficient examples across different handshapes and movements. In contrast, the VGT Corpus \citep{van2015het}, used in prior one-shot research \citep{de2023querying}, follows a Zipfian distribution, where most samples feature the simple pointing handshape. As a result, models trained on VGT struggle to generalise to unseen signs that involve more complex handshapes, as the dataset lacks the necessary variety.
This difference in data diversity explains the significantly higher performance of our \PFASL~model compared to \PFVGT. The PoseFormer model, trained on ASL Citizen, benefits from encountering a wide array of handshapes, enabling it to learn more transferable representations for unknown signs.

%\subsubsection{Pretraining Dataset Variety is More Important than Language}
The variety of the ASL Citizen dataset plays a crucial role in the success of our model in one-shot classification. By ensuring exposure to a wide range of signs, the model minimises the chance of encountering unfamiliar signs, leading to better generalisation. Importantly, our results demonstrate that the diversity within a dataset is more critical than consistency in the language used during pretraining. Even though WLASL \citep{li2020word} and ASL Citizen \citep{desai2024asl} both represent ASL, their partially shared, but differing vocabularies yield markedly different results in pretraining and one-shot classification. This suggests that limiting pretraining to a single dataset or language may not capture the full complexity of an entire sign language, and future models should prioritise dataset variety to enhance robustness in real-world applications.

% subsection about the representation of signs.
Sign representation plays a pivotal role in achieving robust performance in one-shot sign language classification. By focusing on pose-based embeddings, our approach abstracts away from surface-level visual differences, emphasising core elements of sign structure like handshapes, movement, and orientation. This abstraction enables models to generalise across different languages and datasets, which addresses the challenges posed by variations in signer appearance, environment, and video quality. As sign languages evolve and new signs emerge, systems that rely on such representations rather than static translations are better equipped to adapt and remain relevant, paving the way for scalable and inclusive sign language technologies.

% Second, the uniformity of ASL Citizen's class distribution ensures that the model
% is exposed to sufficient examples of a variety of handshapes and movements. This is then in contrast with the VGT Corpus \citep{van2015het} used in prior one-shot sign language recognition research \citep{de2023querying}, for which the class distribution is Zipfian and the majority of examples are for signs that use the simple pointing handshape. The PoseFormer cannot learn
% from such a dataset a robust representation that will transfer to unknown signs, since it will not have
% seen many of the different handshapes that will occur in the set of unknown signs.
% This explains we observe the significantly higher scores with the \PFASL~model,
% when comparing it to the \PFVGT.

\section{Future work}
Despite our considerable gains in ISLR performance and the first results of one-shot ISLR, there are several promising directions for future research. One such direction involves leveraging the lexical information from the ASL-LEX database \citep{caselli2017asl} when using the ASL-citizen dataset. Additionally, extending the existing data with resources like the Sem-Lex benchmark dataset \citep{kezar2023sem} could further enhance recognition performance. However, we argue that the key to achieving more robust sign recognition lies not just in more data but in the diversity of signs it contains. Therefore, multilingual training, where a shared sign representation is used to recognise signs across different languages, may be a more effective approach. Such an approach could vastly expand the potentially recognisable glossary of signs, contributing to more versatile and scalable models.

Finally, the proposed method not only enables highly accurate dictionary search applications but also creates opportunities for other downstream tasks. As mentioned in \cref{sec:oneshot}, \citet{mittal2021compositional} formally described \emph{search}, but also introduced \emph{retrieval} as the extraction of relevant features. We envision the combination of the \emph{search} technique proposed in this paper for the \emph{retrieval} of token embeddings for large language models. The integration into LLMs could facilitate a sign language translation tool or even a sign language-enabled virtual assistant. In summary, due to its effectiveness and lower data requirements, this technique has the potential to significantly advance sign language research, which is currently constrained by data availability. This could lead to the development of more practical tools for the DHH community in the near future.

\section{Conclusion}
Sign language recognition models based on keypoints and self-attention, trained on large-vocabulary
datasets, can classify unknown signs in a different language with just one training example.
We leverage the proven PoseFormer model, pretrain it on the ASL Citizen dataset, and use it in a
one-shot classification setting by leveraging the attention mechanism in the embedding
space of the internal representations learnt by the PoseFormer. The PoseFormer achieves
state-of-the-art sign language recognition on the ASL Citizen dataset (a 19\% increase in Recall@1 compared to previous work)
and on one-shot sign classification (0.508 MRR and 0.374 Recall@1 on 10,235 signs). For the first time, large vocabulary ISLR is enabled thanks to the one-shot classification approach. Furthermore, we prove that the method generalises to different languages and is independent of the used sign-variations inside the support set. Despite the multi-lingual evaluation, we leave the multi-lingual pretraining for one-shot ISLR for future research. Finally, the results of this paper led to the development of a publicly available dictionary look-up application for the DHH community.

% \mathieu{aangezien we enkel over VGT dictionary spreken geeft dit eigenlijk wel weg wie we zijn. willen we trouwens ergens iets zeggen over dat we die one shot datasets ook beschikbaar maken?}
% The method does not yet utilise facial keypoints, which can be important for distinguishing signs based on mouthings. 
% This is left for future research. Also left for future research is the collection of a larger, more representative test set, which is currently in progress.

\bibliography{main}
\bibliographystyle{iclr2025_conference}

\end{document}